\documentclass{iopjournal}
\usepackage{graphicx}
\usepackage{subcaption}
\usepackage{multirow,amsmath,amssymb}

\usepackage[dvipsnames]{xcolor}
\usepackage{dsfont}
\usepackage{siunitx}
\newcommand{\mylabfull}{Labo. de l'Int\'egration du Mat\'eriau au Syst\`eme}
\newcommand{\myorgfull}{Univ. Bordeaux, Bordeaux INP, CNRS}
\newcommand{\myaffilfull}{%
    \mylabfull, \myorgfull - F-33400 Talence, France
}

\begin{document}

\articletype{paper} 

\title{Line-based Event Preprocessing: \\Towards Low-Energy Neuromorphic Computer Vision}

\author{Amélie Gruel$^{1,*}$\orcid{0000-0003-3916-0514}, Pierre Lewden$^1$\orcid{0009-0005-8311-8526}, Adrien F. Vincent$^1$\orcid{0000-0002-9770-4236} and Sylvain Saïghi$^1$\orcid{0000-0002-1414-2523}}

\affil{$^1$\myaffilfull}

\affil{$^*$Author to whom any correspondence should be addressed.}

\email{\{firstname.surname\}@u-bordeaux.fr}


\begin{abstract}
Neuromorphic vision made significant progress in recent years, thanks to the natural match between spiking neural networks and event data in terms of biological inspiration, energy savings, latency and memory use for dynamic visual data processing. However, optimising its energy requirements still remains a challenge within the community, especially for embedded applications.
One solution may reside in preprocessing events to optimise data quantity thus lowering the energy cost on neuromorphic hardware, proportional to the number of synaptic operations. To this end, we extend an end-to-end neuromorphic line detection mechanism to introduce line-based event data preprocessing.
Our results demonstrate on three benchmark event-based datasets that preprocessing leads to an advantageous trade-off between energy consumption and classification performance. Depending on the line-based preprocessing strategy and the complexity of the classification task, we show that one can maintain or increase the classification accuracy while significantly reducing the theoretical energy consumption. Our approach systematically leads to a significant improvement of the neuromorphic classification efficiency, thus laying the groundwork towards a more frugal neuromorphic computer vision thanks to event preprocessing.
\end{abstract}

\section{Introduction}
Computer vision is a thriving field of application for neuromorphic computing, thanks in part to the natural match between Spiking Neural Networks (SNNs)~\cite{Maass_1997} and event-based cameras~\cite{lichtsteiner_128times128_2008}.
Indeed, SNNs aim to implement neural architectures and achieve learning by drawing inspiration from biology more strongly than traditional artificial neural networks (ANNs). To do so, each neuron receives and processes information as spikes, i.e. sequences of electrical pulses, using its membrane potential -- a behaviour mimicking the one of a biological neuron, and leading to a reduced energy consumption when implemented on appropriate hardware. SNNs handle particularly well data produced by event-based cameras~\cite{lichtsteiner_128times128_2008}, a neuromorphic vision sensor inspired by the mechanism of a biological retina: ``events'' are produced asynchronously by each pixel that perceived a significant change in luminosity. 

Combined, these two allow for the efficient implementation of many computer vision tasks~\cite{gallego_eventbased_2020,christensen_2022}. However, significant improvements are still required in terms of energy and memory requirements: due to the event camera's high temporal resolution (at the microsecond scale), a highly dynamic scene recorded by Prophesee's One Megapixel camera~\cite{Finateu2020} for example would produce a substantially large amount of data requiring a large architecture and a high memory demand. Additionally, handling input data of such size as directly captured by the camera would lead to a proportionally larger energy consumption. Indeed, most neuromorphic architectures require the neuron state to be fetched from memory at each synaptic event then be rewritten; each of those operations consumes a certain amount of energy (usually around \qty{10}{\pico\joule})~\cite{sorbaro2019}. Greater input size leads to greater architecture, i.e. a greater number of synapses and higher possibility of synaptic events. 
We thus believe that the preprocessing of event data should be considered before any downstream computer vision task, in order to reduce either or both spatial resolution and number of events for lighter energy requirements. 
\\

\begin{figure*}[ht]
    \centering
    \includegraphics[width=\textwidth]{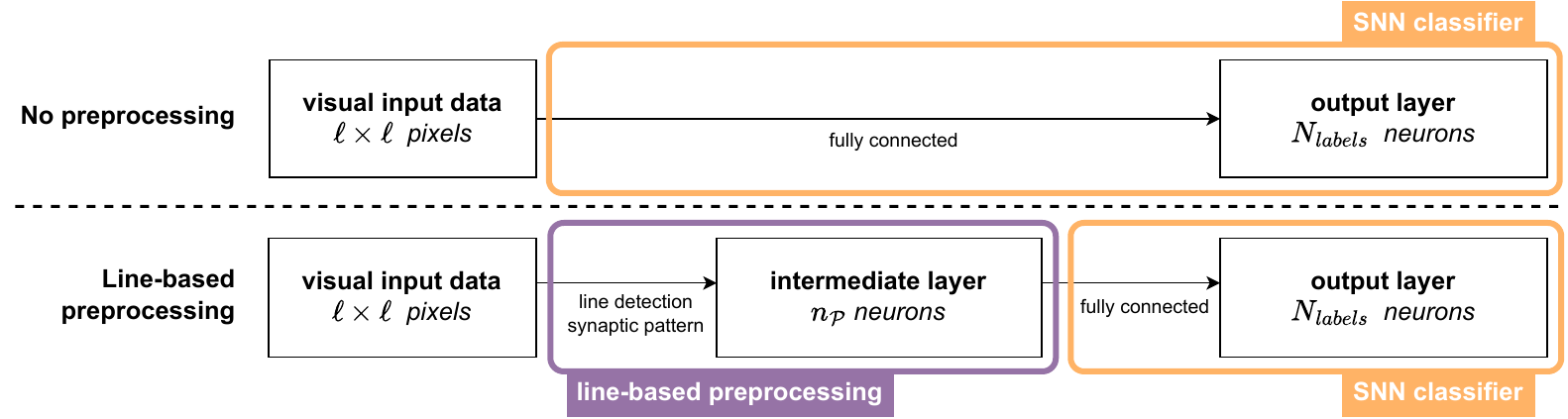}
    \caption{Our protocol for the evaluation of the impact of added line-based preprocessing on neuromorphic classification. The parameters $\ell$ and $n_\mathcal{P}$ respectively stand for the size in pixels of the input sensor and the number of neurons involved in the preprocessing layer (further detailed in Tab.~\ref{tab:nb_neurons_preprocessing}).}
    \label{fig:protocole}
\end{figure*}

Data preprocessing makes for a crucial first step in traditional computer vision or signal processing. Many methods exist that can be combined, leading to diverse results such as reducing noise (using Gaussian blur, Median blur, Laplacian filter, etc.), removing background objects (using edge detection, thresholding, watershed, etc.) or augmenting a dataset (using cropping, rotations, colour manipulations, etc.)~\cite{patel_2023}. Some of those methods have been tentatively adapted to event data in recent works, for example by the Tonic framework~\cite{tonic} --- however, it often requires the accumulation of events into frames and the loss of many advantages of event-based vision. 

One should be mindful to preprocess event data with the aim to preserve, or even improve the neuromorphic energy efficiency and memory usage while retaining relevant information within the data. As discussed earlier, ``reducing the number [of synaptic operations] is the most natural way to keep energy usage low'' in state-of-the-art neuromorphic architectures~\cite{sorbaro2019}. One approach to this challenge could thus be reducing the number of events in the dataset while retaining the information provided, as achieved by the neuromorphic downscaling and foveation developed in~\cite{gruel_2022_visapp,gruel_2023_wacv,gruel_2023_biocyber}. 

A second approach would be to reduce the number of overall synapses, to which the number of synaptic operations is proportionate. It could be achieved by preprocessing event data into differentiable patterns encoded with a smaller number of neurons, leading to a reduction of the number of synapses. This is akin to neuromorphic feature extraction, which aims to ``represent commonly observed spatiotemporal patterns of events [with the] ability to be uniquely distinguished in feature space''~\cite{afshar2020}. Some examples of neuromorphic feature extraction exist in the literature, either based on corner detection~\cite{vasco2016,alzugaray_chli_2018}, edge detection~\cite{ieng2014,brandli2016} or learnt directly from data~\cite{orchard2015,chandrapala2016,lagorce2017,afshar2020}. \\

Since line detection is a fundamental part of perception and comprehension of the visual environment by biological organisms~\cite{kittens,gomez_eguiluz2020}, we propose to adapt the bio-inspired visual mechanism of line detection to the preprocessing of event data as as pictured in Fig.~\ref{fig:protocole}. In this work, we extend our recently published SNN model for line detection within event data~\cite{gruel_aicas_2024}, which is to the best of our knowledge the only existing one to achieve this goal using an end-to-end neuromorphic approach. This previous work obtains a convincing performance with no learning phase and with a minimised architecture, relying solely on intrinsic SNN behaviours and experimentally deployed on SpiNNaker~\cite{spinnaker}.

We apply this line detection model to extract features from event-based benchmark datasets. We introduce different strategies and study their applications to the task of classifying three benchmark datasets varying in ease, data properties and scale: PokerDVS~\cite{PokerDVS}, N-MNIST~\cite{NMNIST} and DVS128 Gesture~\cite{amir2017low}. We estimate the complexity of hardware implementation as well as the memory use thanks to the analysis of the number of neurons and synapses within the architecture used for preprocessing and classification. We evaluate the evolution of the classification performance as well as the theoretical energy efficiency (i.e. the ratio between classification performance and number of synaptic events) with and without line-based preprocessing. 

The focus of this work is on energy efficiency, and its improvement in a traditional neuromorphic computer vision task. We aim to achieve this goal while maintaining a satisfying performance; however improving the latter would only be a secondary benefit and not a mandatory objective to be reached. All in all, the results below validate our objective: we demonstrate that line-based neuromorphic preprocessing allows for a promising trade-off between energy efficiency and performance.

\section{Methods}
\subsection{Neuromorphic model for line detection}
\label{sec:model}

\begin{figure*}[h]
    \centering
    \begin{subfigure}[c]{0.5\textwidth}
        \centering
        \caption{Overall connectivity from input sensor to detectors \\ \textcolor{white}{fill out}}
        \includegraphics[width=\textwidth]{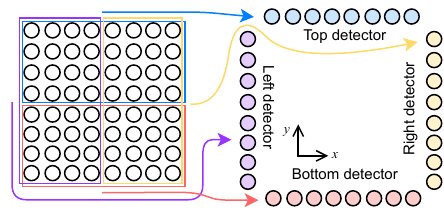}
        \label{fig:line_detection_connectivity}
    \end{subfigure}
    \hfill
    \begin{subfigure}[c]{0.4\textwidth}
        \caption{Details of synaptic patterns}
        \includegraphics[width=\textwidth]{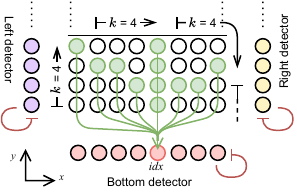}
        \label{fig:line_detection_synapses}
    \end{subfigure}
    \vspace{-0.6cm}
    \caption{Schematic explanation adapted from~\cite{gruel_aicas_2024} of the line detection model used within this work. (a) Overall connectivity from the input sensor (pictured as a square of $8{\times}8$ input pixels) to the four detectors, i.e. the four neuronal populations allowing for preprocessing. (b) Synaptic patterns of activation and inhibition between the input sensor and the bottom detector. In this example, input neurons from the lower half of the sensor are connected to the neuron $idx$ in the bottom detector with a step of $k=4$ between the diagonal connections and with an activating strength $\omega$ (see details in Sec.~\ref{sec:hyperparam_finetune}). This pattern is reproduced for all neurons in all four detectors. Winner-Takes-All is applied on each detector to ensure model selectivity.}
    \label{fig:line_detection}
\end{figure*}

The line detection model we introduced in~\cite{gruel_aicas_2024} detects static or moving lines in the input event data spanning the whole sensor by outputting the position of their intersections with the sensor's borders. For each line in the input event data, two detectors activate corresponding to the edges crossed by the input line within ``top'', ``bottom'', ``right'' and ``left'' halves (see Fig.~\ref{fig:line_detection_connectivity}). From this activity, one can then extrapolate the coordinates of the line extrema, thus the position and orientation of the whole line. 

This mechanism is implemented using one output layer of Leaky-Integrate-and-Fire (LIF) neurons~\cite{paugammoisy} comprising four parallel populations of neurons, each corresponding to one border of the sensor and named accordingly. Those neural populations will be referred to as ``detectors'' for the remainder of this work, as opposed to ``sensor'' which describes the DVS camera providing input events. The leakiness of the neurons allows for stable detection of a moving line over time~\cite{gruel_aicas_2024}. The selectivity of the model is ensured thanks to a Winner-Takes-All mechanism applied to each detector \textit{via} lateral inhibition within the detectors (see Fig.~\ref{fig:line_detection_synapses}).

Each neuron of each detector receives activations from input neurons situated on diagonals spanning over the corresponding half-sensor. Diagonal connections are implemented at every $k$ potential position, with $k = 1$ for a pattern encompassing all potential diagonals. Fig.~\ref{fig:line_detection_synapses} schematises the synaptic pattern for $k=4$ from the bottom-half of the input sensor to the neuron $idx$ within the bottom detector. The higher the value $k$ is set to, the lower the number of diagonals in the synaptic pattern is, thus the lower the number of synapses to implement is and the lighter the architecture becomes. 

\subsection{Adaptation to event data preprocessing}

\begin{figure*}[h]
    \centering
    \begin{subfigure}[c]{\textwidth}
        \caption{Different strategies for line-based preprocessing}
        \label{fig:detectors_preprocess}
        \includegraphics[width=\textwidth]{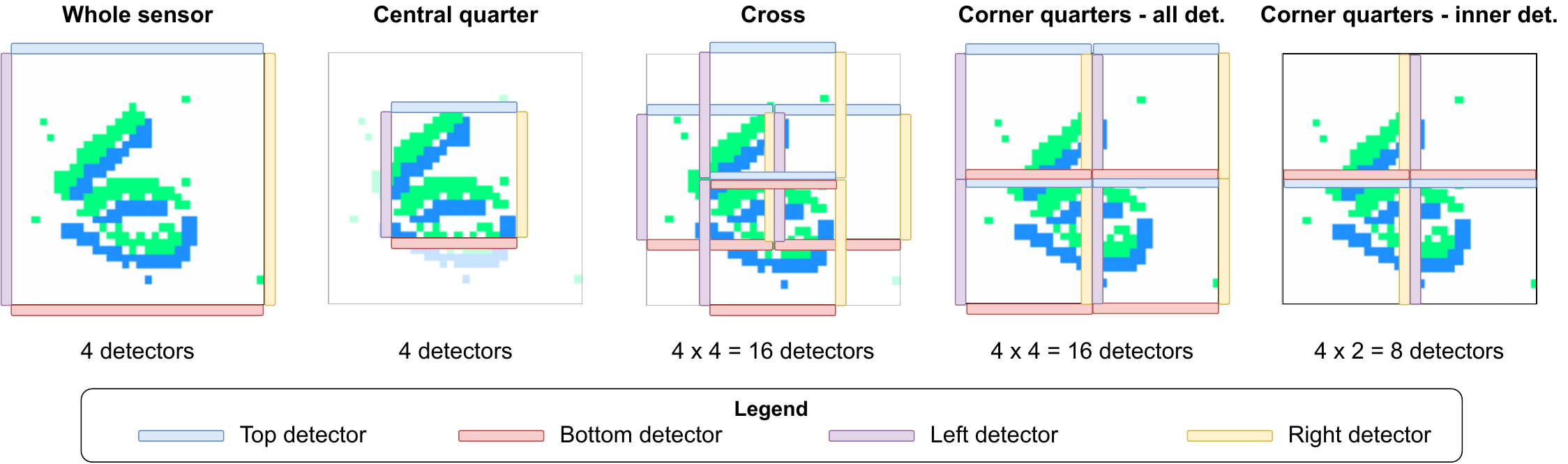}
    \end{subfigure} \\
    \begin{subfigure}[c]{0.48\textwidth}
        \centering
        \caption{Input events and corresponding spikes emitted by the ``Corner quarters -- inner detectors'' strategy.}
        \label{fig:input_output_NMNIST_6_2d_split}
        \includegraphics[trim={0 0.3cm 0 0},clip,width=\textwidth]{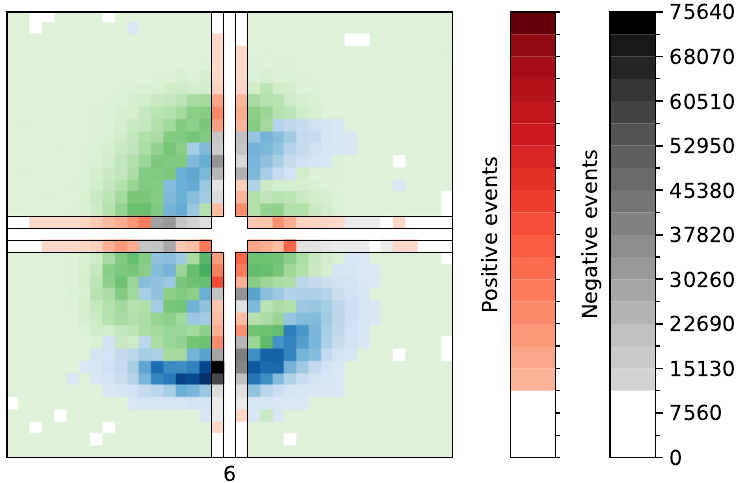}
    \end{subfigure}
    \hfill
    \begin{subfigure}[c]{0.49\textwidth}
        \centering
        \caption{Aggregated spikes emitted by the intermediate ``Corner quarters -- inner detectors'' strategy.}
        \label{fig:output_NMNIST_6_2d_split}
        \includegraphics[trim={0 1cm 2.7cm 1cm},clip,width=\textwidth]{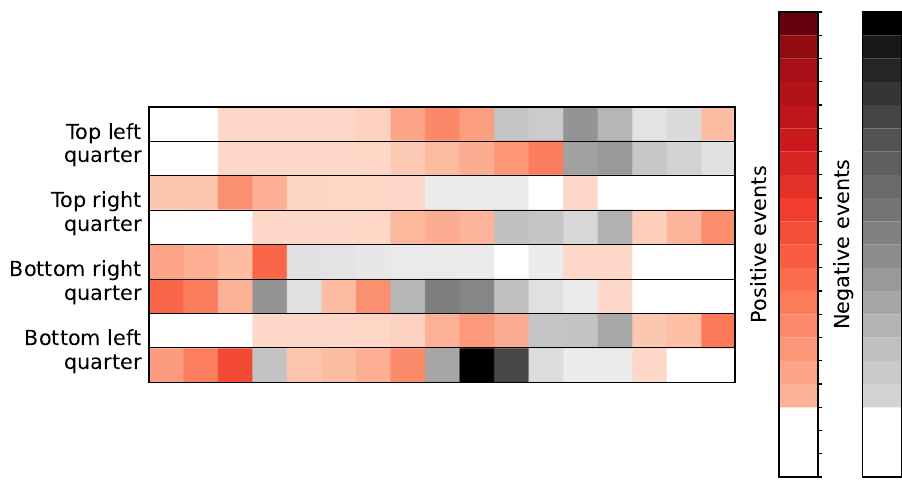}
    \end{subfigure}
    \caption{(a) Schematic representation of the different proposed strategies for line-based preprocessing, applied to an input N-MNIST sample as an example. (b) Input and output spikes obtained from the strategy ``corner quarter - inner detectors'' with split polarities averaged on N-MNIST samples labeled as 6. Green and blue dots correspond respectively to the positive and negative input events; red and black correspond respectively to the model outputs for positive and negative polarities. Events are averaged over 7,000 ``train'' and ``test'' samples. (c) Output spikes produced by the intermediate preprocessing layer in the same context as (b), aggregated in a 2D format.}
    \label{fig:line_detection_all}
\end{figure*}

In this paper, we propose a novel approach to event data preprocessing, by extracting linear features within input events using the model from~\cite{gruel_aicas_2024} described previously. As pictured in Fig.~\ref{fig:detectors_preprocess} and further detailed in Tab.~\ref{tab:nb_neurons_preprocessing}, for an input sensor of size $\ell {\times} \ell$ pixels (with $\ell$ the length of the sensor in pixels), we consider five different strategies:
\begin{itemize}
    \item ``Whole sensor'': the line detection mechanism is applied as is to the entirety of the input sensor. The output is spread over 4 detectors, thus comprises $4 \ell$ neurons. 
    \item ``Central quarter'': the line detection is applied to a central patch of size $\ell / 2 \times \ell / 2$ pixels, designated as ``quarter'' in the remainder of this paper. This second approach dismisses the information provided by pixels outside the central quarter and relies on 4 detectors of halved dimension, thus on $2 \ell$ output neurons.
    \item ``Cross'': the input sensor is split into four quarters forming a cross. The quarters overlap within the central part of the sensor, and data emitted in the sensor corners are dismissed. The line detection model is applied as is independently to each quarter, resulting in 16 detectors and $8 \ell$ output neurons. 
    \item ``Corner quarters - all detectors'': this strategy varies from the previous one by the spatial distribution of the four quarters. Each is positioned in a different corner of the sensor, so that no event is dismissed and there is no overlap in the central part. It also involves 16 detectors and $8 \ell$ output neurons. 
    \item ``Corner quarter - inner detectors'': this strategy is similar to the previous one --- however, instead of keeping 4 detectors per quarter, only the 2 inner detectors are implemented (namely the top and left detector for the quarter on the bottom right, the top and right detector for the quarter on the bottom left, the bottom and left detector for the quarter on the top right and the bottom and right detector for the quarter on the top left). Indeed, the events carried by certain neuromorphic datasets such as Poker-DVS~\cite{PokerDVS}, N-MNIST~\cite{NMNIST}, CIFAR10-DVS~\cite{CIFAR10} or DVS128 Gesture~\cite{amir2017low} are mostly clustered in the sensor centre thus leading to a quasi-null activity in outer detectors in the previous strategy. One should note that reducing the number of detectors (from 16 to 8 detectors) leads to a reduction of neurons (from $8 \ell$ to $4 \ell$ output neurons) and synapses, i.e. a reduction of memory use and energy consumption. 
\end{itemize}

\begin{table}[h]
    \centering
    \caption{Number of neurons within the preprocessing layer depending on the adopted strategy.}
    \begin{tabular}{c c}
        \hline
        \textbf{Strategy} & \textbf{Number of preprocessing neurons $n_\mathcal{P}$} \\
        \hline
        Whole sensor & $4 \ell$ \\
        Central quarter & $4 \times \ell/2 = 2 \ell$ \\
        Cross & $4 \times 4 \times \ell/2 = 8 \ell$ \\
        Corner quarters -- all detectors & $4 \times 4 \times \ell/2 = 8 \ell$ \\
        Corner quarters -- inner detectors & $4 \times 2 \times \ell/2 = 4 \ell$ \\
        \hline
    \end{tabular}
    \label{tab:nb_neurons_preprocessing}
\end{table}

Each strategy has two variants: either it considers all input events similarly with no regard to polarity (``merged polarities''), or it distinguishes positive and negative events and applies line detection per polarity (``split polarities''). In most existing SNN simulators, event polarity cannot be directly translated into positive and negative spikes; to achieve the second variant, one must split the events into two neuronal populations (one per polarity). The application of the ``split polarities'' variant within our line-based event preprocessing mechanism thus doubles the number of neurons in the input layer as well as the number of synapses. Therefore we consider each variant for each strategy separately. 

The line-based preprocessing step is implemented on CPU using the NEST simulator~\cite{gewaltig_nest_2007} interfaced with PyNN~\cite{davison2009}. The neuronal and synaptic parameters used for this implementation are detailed in Tab.~\ref{tab:parameters}. To be noted that the LIF neurons are implemented using PyNN's object \texttt{IF\_cond\_exp}, which corresponds to a LIF with an exponentially-decaying post-synaptic conductance.

\begin{table}[hb]
    \centering \small
    \caption{Neuronal and synaptic parameters involved in the line-based preprocessing step.}
    \begin{tabular}{c c}
        \hline
        \textbf{Parameters} & \textbf{Values}  \\
        \hline
        Resting membrane potential $v_{rest}$ & \qty{-60}{\milli\volt} \\
        Reset membrane potential $v_{reset}$ & \qty{-60}{\milli\volt} \\
        Neuronal threshold $\theta$ & \qty{-30}{\milli\volt} \\
        Refractory period $\tau_{r}$ & \qty{0.1}{\milli\second} \\
        Membrane time constant $\tau_m$ & \qty{2.5}{\milli\second} \\
        Winner-Takes-All inhibition weight $\omega_{WTA}$ & 1 \\
        Excitatory and inhibitory delay $\delta$ & 1 \\
        Decay time of the synaptic conductance $\tau_{syn}$ & \qty{5}{\milli\second} \\
        \hline 
    \end{tabular}
    \label{tab:parameters}
\end{table}

\subsection{Experimental protocol}

In the remainder of this work, we will discuss the pros and cons of line-based event data preprocessing in terms of energy consumption and classification accuracy, compared to the one obtained with no preprocessing phase. Designated as the ``no preprocessing'' strategy in the remainder of this paper, the latter corresponds to directly providing the input events to a fully connected SNN classifier with no hidden layer. As for the five line-based preprocessing strategies, we will consider the preprocessing phase as integrated into the classification pipeline in the discussion of our results: the input events are provided to the detectors according to the line detection synaptic patterns, then the output of the detectors' neurons are fed into the classifier \textit{via} a fully-connected synaptic pattern (see Fig.~\ref{fig:protocole}). 

\paragraph{Hyperparameters fine-tuning}
\label{sec:hyperparam_finetune}
In order to obtain the optimal configuration of the five line-based preprocessing approaches, two hyperparameters of the preprocessing layer are fine-tuned for each variant: 
\begin{itemize}
    \item the step $k$ between diagonals within the line detection model, introduced in Sec.~\ref{sec:model}, is assessed for values ranging from 1, 5, 10, ... to 30. Depending on the dataset under study, input sensor size varies from $34 {\times} 34$ to $128 {\times} 128$ (see Tab.~\ref{tab:datasets}).
    \item the strength $\omega$, i.e. the sum of the activations received by a neuron within the preprocess layer from the input layer according to the diagonal synaptic pattern. Its calculation is detailed in~\cite{gruel_aicas_2024}. The strength $\omega$ is fine-tuned among values ranging in $\{1;2.5;5;7.5;10\}$.
\end{itemize}
It is expected that the higher $k$ is set to, the lower the number of synaptic events within the classifier is, as a higher step between synapses leads to fewer synapses, thus fewer synaptic events. However a high $k$ also induces a lower precision in the detected lines. Similarly, a higher $\omega$ will lead to a stronger activation, i.e. a higher number of emitted spikes. For each preprocessing strategy, two sets of hyperparameters are identified as optimal:
\begin{itemize}
    \item the couple $(k,\omega)$ allowing for best inference accuracy, referred to as the ``best accuracy parametrisation'' in the remainder ;
    \item a second couple $(k,\omega)$ allowing this time for best inference efficiency (i.e. the ratio of accuracy by the number of synaptic events), referred to as the ``best efficiency parametrisation'' in the remainder.
\end{itemize}

\paragraph{Classifier}
\label{sec:classifier}
The classifier used in this work is voluntarily simple, as we wish to study the impact of line-based preprocessing on neuromorphic efficiency within a simple context. Indeed, we take no interest in improving overall classification accuracy on benchmark datasets, but we aim to improve energy efficiency in a traditional computer vision task, especially for dedicated embedded hardware platforms. As stated above, the classifier is composed of a single fully-connected (i.e. linear) layer. It takes as input events either after preprocessing or directly as is, and its output layer is composed of $N_{labels}$ LIF neurons where $N_{labels}$ corresponds to the number of labels within the dataset. The classifier is trained over 10 epochs using backpropagation through time relying on the Adam optimiser~\cite{adamoptimiser} (initiated with a learning rate of 0.03) and on a mean-square-error spike-count loss. The latter allows for rate-base decoding of the spiking activity, where the output neuron emitting the most spikes in a given time indicates the detected label, while limiting the dead neuron problem by encouraging output activity even from incorrect neurons. The LIF neurons are implemented as follows using the SNNtorch framework~\cite{snntorch}: they use the arctangent surrogate gradient function adapted from~\cite{Fang2020IncorporatingLM}; their membrane potential, which decay rate $\beta$ is set to 0.9, is reset by subtraction at each emitted spike. The threshold of the LIF neurons and the batch size vary according to the dataset (see Tab.~\ref{tab:datasets}). All results presented below are averaged over three runs with varying seeds.

\begin{table}[ht]
    \caption{Datasets under study and corresponding classifier parameters.}
    \centering
    \begin{tabular}{l c c c}
        \hline
        \textbf{Dataset} & \textbf{PokerDVS~\cite{PokerDVS}} & \textbf{N-MNIST~\cite{NMNIST}} & \textbf{DVS128 Gesture~\cite{amir2017low}} \\
        \hline
        Sensor size (in pixels) & $35 {\times} 35$ & $34 {\times} 34$ & $128 {\times} 128$ \\
        $N_{labels}$ & 4 & 10 & 11 \\
        Samples distribution (train/test) & 48/12 & 60,000/10,000 & 1078/264 \\
        Original, averaged sample size & \qty{17}{\milli\second} & \qty{310}{\milli\second} & \qty{6.5}{\second} \\
        Shortened sample size used & \qty{10}{\milli\second} & \qty{10}{\milli\second} & \qty{1}{\second} \\
        Batch size & 8 & 128 & 128 \\ 
        LIF threshold & 5 & 25 & 15 \\
        \hline
    \end{tabular}
    \label{tab:datasets}
\end{table}

\paragraph{Input data}
Matching available examples of similar classifiers, preliminary event transformations provided by the Tonic framework~\cite{tonic} are applied to the input events regardless of the preprocessing (or lack of). Each sample is denoised and its events are accumulated into frames over \qty{1}{\milli\second}. Samples are shortened to smaller temporal size provided in Tab.~\ref{tab:datasets}. Such preprocessed events lead to satisfying classification performance while significantly reducing the computation time of the current study. 

\paragraph{Evaluation metrics}
As stated above, the goal of this work is to assess the benefits of line-based preprocessing on the energy efficiency and performance of neuromorphic computer vision. We compare different strategies according to three main metrics: the classification accuracy $\mathds{A}$, the number of synaptic events $SE$ and the theoretical efficiency $\mathds{E}$ at inference. 

The accuracy corresponds to the number of correct inference guesses over the total number of guesses performed by the classifier once the 10-epoch training phase is over. \newline

Building on existing works~\cite{dampfhoffer2023}, one can approximate the theoretical energy consumption of an SNN by the number of synaptic events $SE$ propagating within it during inference. Indeed, many studies show that the energy consumed by an SNN model is proportional to the number of synaptic events in most state-of-the-art neuromorphic architectures~\cite{sorbaro2019}.
In our model architecture and with preprocessing applied, the number of synaptic events $SE$ is calculated as follows: 
\begin{equation}
    \label{eq:n_synaptic_ev_preprocessing} 
    SE = SE_{\mathcal{P}}(k) + SE_{\mathcal{C}} \hspace{0.5cm}\text{with}\hspace{0.5cm} SE_{\mathcal{C}} = \mathcal{E}_{\mathcal{P}} \times n_{\mathcal{C}}
\end{equation}
$\mathcal{P}$ and $\mathcal{C}$ designate respectively the preprocessing and classifier steps, with $\mathcal{E}_{\mathcal{P}}$ the average number of events per preprocessed sample. $n_{\mathcal{C}}$ is the number of output neurons in the classifier, i.e. the number of labels $N_{labels}$ to be classified (see Tab.~\ref{tab:datasets}). 
$SE_{\mathcal{P}}(k)$ is the grand sum of the element-wise product of the average pixel activation in the input dataset per the number of outgoing synapses per pixel. It is difficult to obtain other than experimentally as it highly varies depending on the input data size, the preprocessing strategy adopted and the chosen parameter $k$ (i.e. the step between diagonals involved in line detection). 

Secondly, the number of synaptic events for the ``no preprocessing'' scenario (i.e. only direct classification of event data as is) is calculated as follows, where $\mathcal{E}$ corresponds to the average number of events per sample within the input event dataset: 
\begin{equation}
    \label{eq:n_synaptic_ev_original}
    SE = SE_{\mathcal{C}} = \mathcal{E} \times n_{\mathcal{C}}
\end{equation}

Finally, we define the efficiency $\mathds{E}$ of an SNN as the ratio of the inference accuracy and the corresponding number of synaptic events within the whole network : 
\begin{equation}
    \label{eq:efficiency}
    \mathds{E} = \frac{\mathds{A}}{SE}
\end{equation}
To be noted that we only consider the efficiency of models leading to a satisfying accuracy $\mathds{A}_\mathds{E}$: indeed, a model might reach an impressive efficiency while performing very poorly and not recognising anything, due to a negligible firing activity. To achieve this, we arbitrarily set a threshold $\mathds{A}_{\text{T}}$, bellow which the efficiency is discarded:  
\begin{equation}
    \label{eq:threshold}
    \mathds{A}_{\mathds{E}} \ge \mathds{A}_{\text{T}} \hspace{0.5cm}\text{where}\hspace{0.5cm} \mathds{A}_{\text{T}} = \frac{2}{3} \times \mathds{A}_{\text{no preprocessing}}
\end{equation}

The higher the accuracy and the efficiency are, the better the overall performance. The goal of this work is to reach a significantly higher inference efficiency with similarly satisfying accuracy on benchmark neuromorphic datasets.

\section{Analysis of the model's architecture}
\label{sec:architecture}

This section studies the complexity of the SNN model involved in the protocol described above, according to the adopted strategies (see the six possible approaches in Fig.~\ref{fig:detectors_preprocess}) and in terms of number of neurons and synapses. We assess the different variables according to an increasing sensor size, which values approximate those found in the state-of-the art: common benchmark datasets such as N-MNIST~\cite{NMNIST} or PokerDVS~\cite{PokerDVS} are recorded with a camera size approaching $30 {\times} 30$ pixels; and as stated in their name, the sizes of the cameras involved in the creation of DVS128 Gesture~\cite{amir2017low} and Prophesee 1 Megapixel Automotive Detection Dataset~\cite{Perot2020} reach respectively $128 {\times} 128$ and $1000 {\times} 1000$ pixels.

\begin{figure}[ht]
    \centering
    \includegraphics[width=\textwidth]{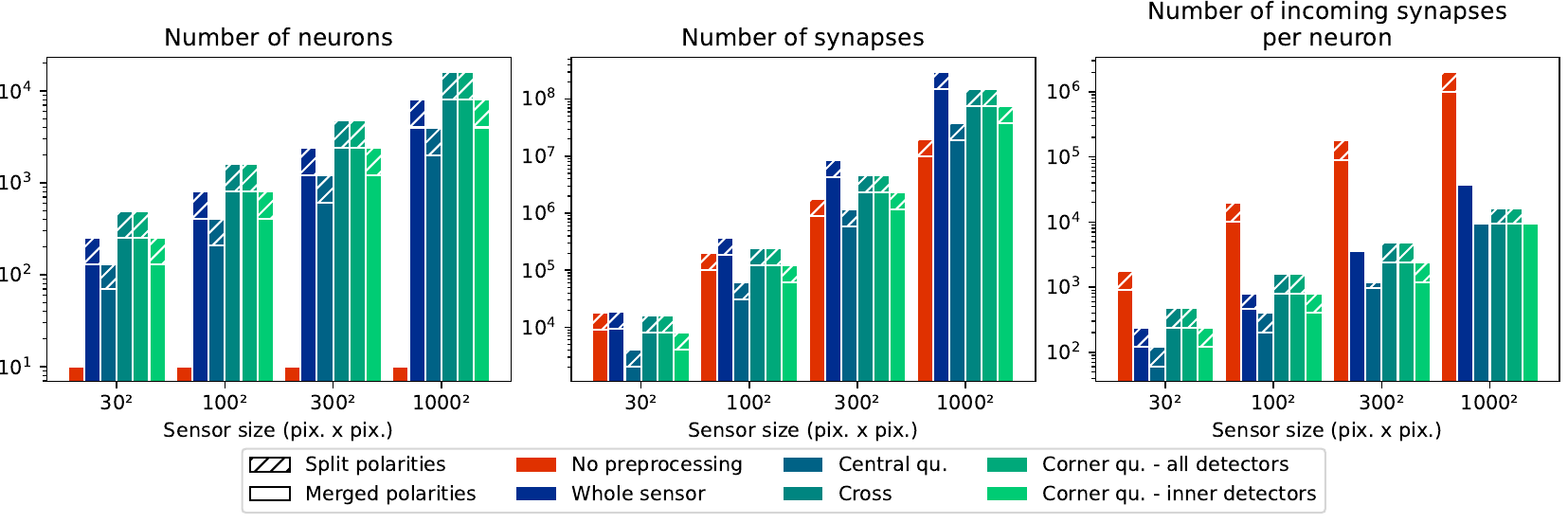}
    \caption{Model architecture depending on the sensor size. The model is characterised by the number of neurons (left plot) and synapses (center plot) involved, as well as by the number of incoming synapses per neuron (right). The number of synapses is computed for a step $k=30$. The bars for merged (no hatches) and split polarities (diagonal hatches) approaches overlap: the split polarities systematically leads to an equal or higher value than the merged polarities one.}
    \label{fig:architecture}
\end{figure}

Most neuromorphic platforms (such as SpiNNaker~\cite{spinnaker} or Loihi~\cite{loihi}) implement the input layer as simple ``encoding neurons'', where each neuron correspond to a pixel (in the case of visual input data) and directly emits spikes according to the AER data as input. These encoding neurons do not require the complex dynamics of a LIF neuron, and are therefore not taken into account in the census of neurons involved in the model. Fig.~\ref{fig:architecture} (left) presents the number of neurons involved within the whole classification pipeline, with or without line-based preprocessing, i.e. sums the number of neurons within the preprocessing layer to those in the output layer. The number of neurons is thus evidently lower within the ``no preprocessing'' approach. However, most neuromorphic platforms support time-division multiplexing for neuronal computation~\cite{SpinLink2012,LoihiTDM2019,ma2023darwin3}: a high number of neurons can be simulated with no significant additional cost in terms of hardware resources or energy consumption. 

The main limit in hardware implementation mostly relies in the number of synapses. As can be observed in Fig.~\ref{fig:architecture} (middle), the overall number of synapses is closer among different approaches than the number of neurons. Since time-division multiplexing also applies itself to the synaptic connections among neurons, it is far more relevant to consider the number of incoming synapses per neuron instead of the global number of synapses involved in the model. Indeed, the hardware implementation of synapses connecting to a spiking neuron is often limited by the available resources; for example, SpiNNaker boards have an upper bound of $1,000$ incoming synapses per neuron~\cite{spinnaker}. According to Fig.~\ref{fig:architecture} (right), the baseline approach of fully-connecting the input sensor to the output layer requires a high number of incoming synapses, which increases quadratically with the sensor size. Adding an intermediate preprocessing layer, such as is the case within the line-based preprocessing method, significantly reduces the number of incoming synapses by multiple orders of magnitude.

As expected, the number of neurons and synapses are identical between the ``cross'' and the ``corner quarters -- all detectors'' strategies, as both relies on the same number of detectors of same size. The lighter architecture corresponds to the ``central quarter'' strategy, which only requires four (half-length) detectors. One can expect that this strategy will be the most efficient one due to its smaller number of synapses, however it is likely that the extraction of features only within the sensor's central quarter will lead to a poor classification accuracy.

\section{Proof of concept on a toy dataset}
We first apply our different line-based preprocessing to PokerDVS~\cite{PokerDVS}, a natively neuromorphic dataset for visual classification of poker symbols. It consists in 58 samples corresponding to one of four possible pips among ``heart'', ``spade'', ``diamond'' and ``club''. It was recorded in 2013 using a $128 \times 128$ DVS camera mainly as a proof of the DVS high-rate recording capacities. As indicated in Tab.~\ref{tab:datasets}, each sample pip was extracted from the original recording and the corresponding samples were spatially resized to $35 \times 35$ pixels. 

\subsection{Hyperparameter fine-tuning}

\begin{figure}[ht]
    \centering
    \begin{subfigure}[c]{\textwidth}
        \caption{Merged polarities}
        \label{fig:hyperparam_PokerDVS_merged}
        \includegraphics[width=\textwidth]{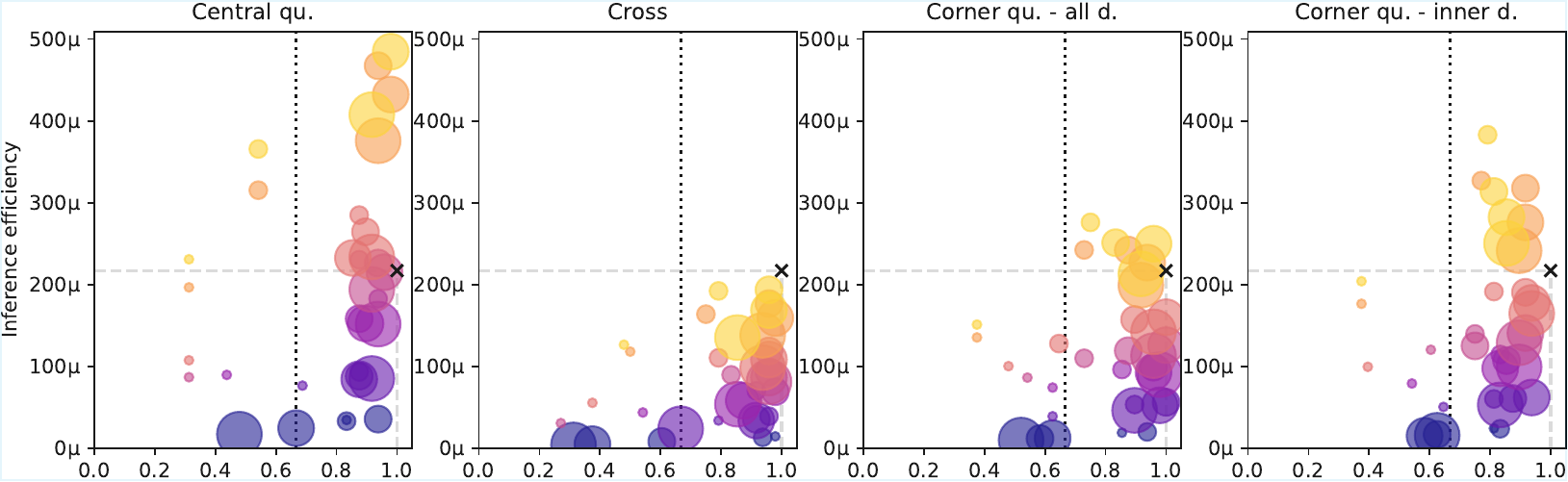}
    \end{subfigure} \\
    \begin{subfigure}[c]{\textwidth}
        \caption{Split polarities}
        \label{fig:hyperparam_PokerDVS_split}
        \includegraphics[width=\textwidth]{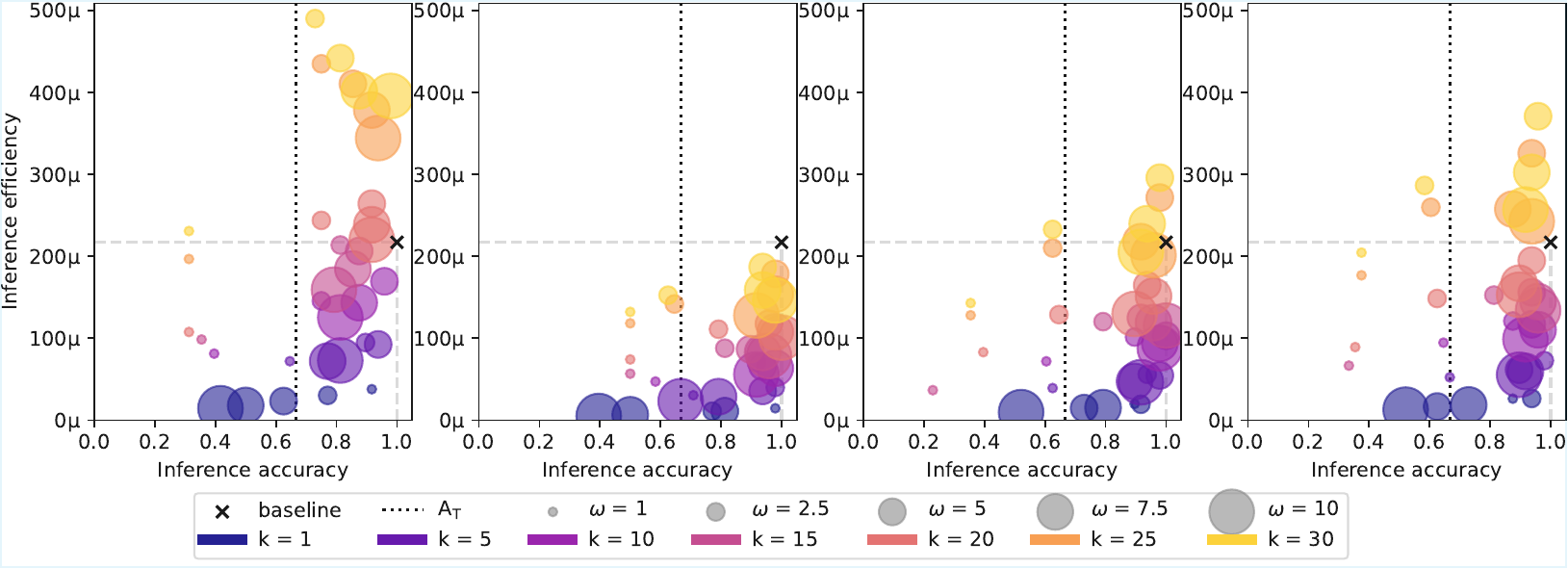}
    \end{subfigure} \\
    \caption{Trade-off between inference efficiency $\mathds{E}$ (y-axis) and inference accuracy $\mathds{A}$ (x-axis) for varying set of hyperparameters $(k,\omega)$ according to the strategies for different line-based preprocessing applied to PokerDVS. (a) and (b) correspond respectively to merged and split polarities, with a shared legend. The ``no preprocessing'' trade-off is indicated by a black cross. The variation of the hyperparameter $k$ are indicated by the colour variation, whereas the size of each point increases with the strength $\omega$. The dotted black line indicates the $\mathds{A}_\text{T}$ threshold.} 
    \label{fig:hyperparam_PokerDVS}
\end{figure}

\begin{figure}[ht]
    \centering
    \includegraphics[width=0.7\textwidth]{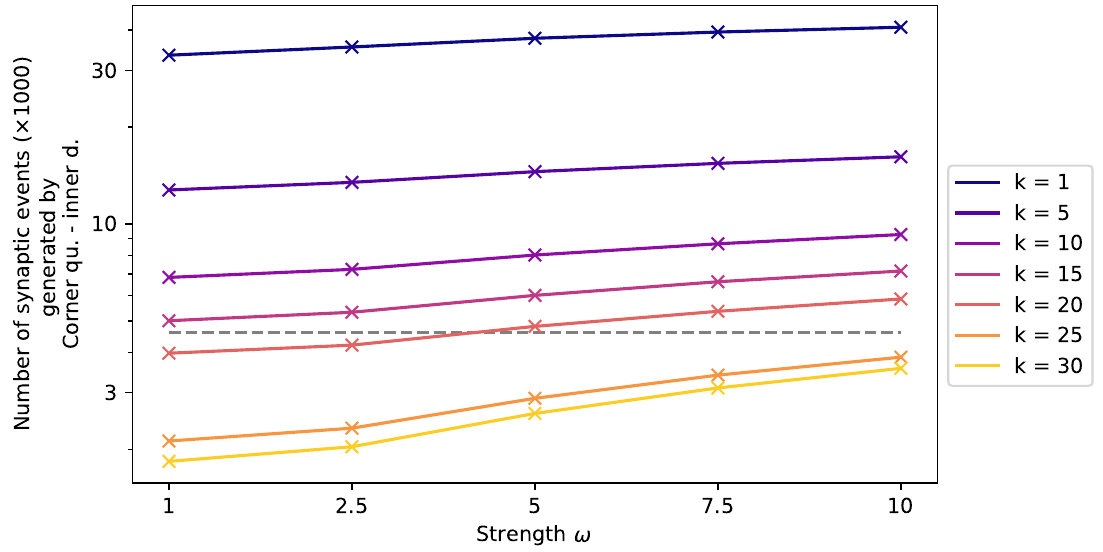}
    \caption{Number of synaptic events produced within the whole pipeline, when applying the ``corner quarters -- inner detectors'' strategy with split polarities to PokerDVS. The colours denote the variation of the step $k$, while the x-axis indicates an increasing strength $\omega$. The dotted black line indicates the baseline number of synaptic events produced by the ``no preprocessing'' approach. Note that the y-axis follows a logarithmic graduation. }
    \label{fig:nbevsyn_PokerDVS}
\end{figure}

We chose this simple dataset, which neuromorphic classification \textit{via} training with backpropagation through time over 10 epochs easily reaches 100\% accuracy, as a first support for validating our line-based preprocessing concept. The results of the hyperparameters fine-tuning phase described in Sec.~\ref{sec:hyperparam_finetune} are presented in Fig.~\ref{fig:hyperparam_PokerDVS}: the trade-off between inference accuracy and inference efficiency is represented for each strategy, with polarities either merged (Fig.~\ref{fig:hyperparam_PokerDVS_merged}) or considered separately (Fig.~\ref{fig:hyperparam_PokerDVS_split}). Our goal is to overcome the inference efficiency while maintaining the accuracy. In other words, we wish to obtain a trade-off above the horizontal dotted gray line (indicating the baseline efficiency attained by the ``No preprocessing'' approach) and laterally as close as possible to the right side of the graph, i.e. to the perfect classification accuracy obtained by the baseline. 

According to  Fig.~\ref{fig:hyperparam_PokerDVS}, many hyperparameter sets reach this targeted top-right corner for most strategies: we can specifically highlight the high efficiency reached as expected by the ``central quarter'' strategy, but also the stable accuracy paired with a significantly high efficiency obtained by multiple sets within the ``corner quarters -- inner detectors'' strategy. We can also observe that higher values of $k$ allow for better efficiencies while conserving good accuracies. This is confirmed by a further study depicted in Fig.~\ref{fig:nbevsyn_PokerDVS}, which represents the number of synaptic events within the whole pipeline for the ``corner quarters -- inner detectors'' strategy with split polarities applied to PokerDVS. As previously intuited, the number of synapses connecting the input to the line-based preprocessing layer, directly dependant to the step $k$, influences greatly the number of synaptic events within the whole pipeline (preprocessing step and classifier combined): the higher the value of $k$, the lower the number of synapses and the lower the number of synaptic events. For a value of $k$ greater than 20, the number of synaptic events and the corresponding theoretical energy consumption are lower than the ones obtained with the baseline approach, despite the additional layer. 

It is also interesting to note that lower values of $k$ distances the trade-off both from a good efficiency as well as a good accuracy. We initially hypothesised the opposite, as a small $k$ leads to a finer steps between the lines detected by our model, \textit{a priori} carrying finer details for the classification (see Fig.~\ref{fig:line_detection_all}). This counter-intuitive observation should be compared with results obtained in a more complex context than the one provided by the classification of PokerDVS, an all-in-all quite easily solvable task.

\subsection{Comparison with similar approaches}
\begin{figure}[ht]
    \centering
    \includegraphics[width=\textwidth]{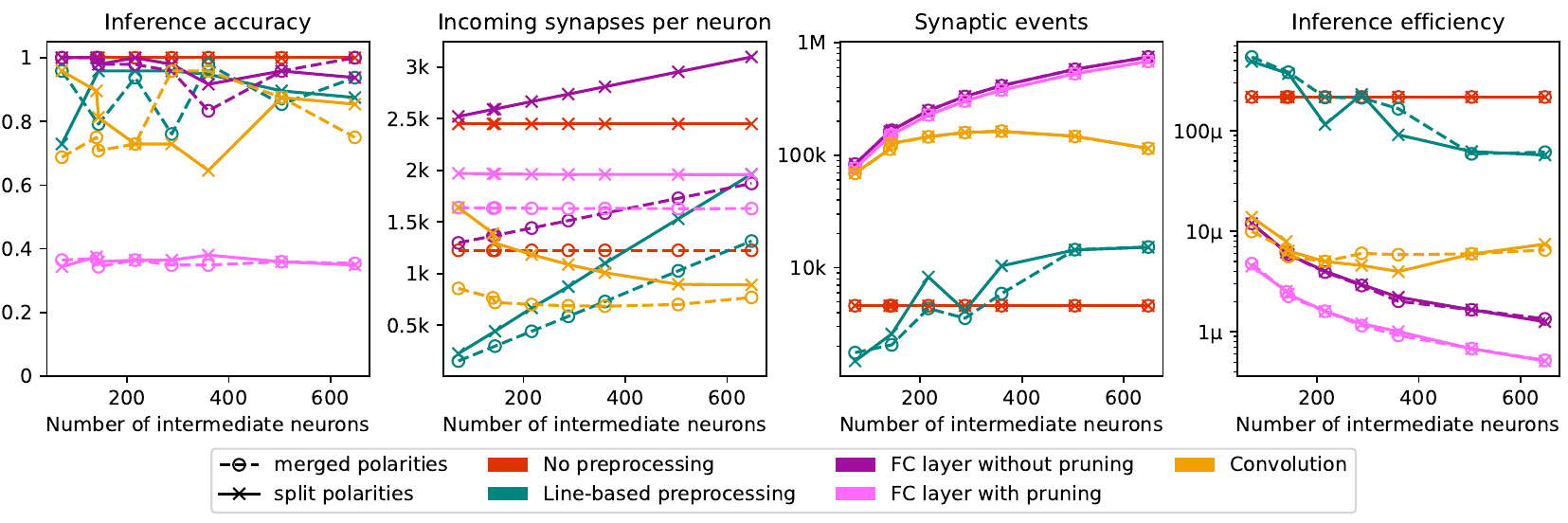}
    \caption{Performance comparison between the ``No-preprocessing'' (red lines) and various line-based preprocessing strategies (turquoise lines), with a convolution approach (yellow lines) and a fully connected one (``FC layer''), before (purple lines) or after (pink lines) pruning on the first layer. The last three approaches use an intermediate hidden layer, which number of neurons (x-axis) varies according to the number of neurons involved in the preprocessing layer for the five line-based preprocessing strategies introduced in this paper.}
    \label{fig:compar_PokerDVS}
\end{figure}

As previously stated, the line-based preprocessing method introduced here is opposed to the ``no preprocessing'' approach throughout this work. However, other network architectures and methods commonly used in neuromorphic vision could be assimilated to intermediate preprocessing allowing for feature extraction. Results presented in Fig.~\ref{fig:compar_PokerDVS} help us assess the scientific added-value of comparing our approach to two of such techniques, namely the convolution and the use of an intermediate hidden layer with and without pruning. 

Both the convolution and the pruning approaches are implemented using the classifier parametrisation described in Sec.~\ref{sec:classifier}, and results are computed with event polarities both merged and split. The convolution approach consists in a first 2D convolutional layer, feeding the input events to an intermediate hidden layer composed of $n_i$ LIF neurons, itself fully connected to the output layer. The convolution is applied with a stride of 1, a null padding and a kernel size calculated so that the number of neurons $n_i$ equals the number of neurons in the preprocessing layer in our line-based preprocessing approach. The weights are learned over the whole network, and the bias solely in the second, fully-connected layer. 

Similarly, the fully-connected approach (referenced as ``FC layer'' in the remainder of this work) is composed of one hidden layer composed of $n_i$ LIF neurons, receiving and emitting spikes \textit{via} two fully-connected layers with learnt weights and biases. Those pretrained parameters are then used to obtain the inference performance as is (``FC layer without pruning'' in Fig.~\ref{fig:compar_PokerDVS}) and after a pruning phase (``FC layer with pruning'' in Fig.~\ref{fig:compar_PokerDVS}). The implementation of the latter consists in setting to zero the pretrained synaptic weights of absolute value inferior to a certain threshold. The pruning threshold is empirically set to \qty{\pm 0.01}{}. 

Fig.~\ref{fig:compar_PokerDVS} presents the evolution of the inference accuracy and efficiency, as well as the number of incoming synapses per neuron (see Sec.~\ref{sec:architecture} for further discussion on time-division multiplexing) and the number of synaptic events emitted throughout each network, according to $n_i$. To achieve our goal of lower theoretical energy consumption, the best case scenario reaches a high accuracy and efficiency, combined with a low number of incoming synapses per neuron and number of synaptic events. Both convolution and ``FC layer without pruning'' reach similar or higher accuracy as the line-based preprocessing approach, approaching the ideal performance reached by the ``no preprocessing''. However, adding pruning to the FC layer lowers significantly the inference accuracy, nearing chance-level; it seems that the classification performance of FC layer is sensible to fine details detected by low impact connections even through the number of synaptic events remains similar before and after pruning (see Fig.~\ref{fig:compar_PokerDVS}, third plot). We could have evaluated the FC layer approach with deeper networks, but we can easily predict from our results that increasing the depth would increase the synaptic activity, thus proportionally reducing the energy efficiency. 

Focusing on high-performance approaches: for an increasing amount $n_i$ of intermediate neurons, ``FC layer without pruning'' behaves similarly to the line-based preprocessing and has an increasing, albeit constantly superior, number of incoming synapses per neuron; convolution however presents the opposite behaviour and its number of incoming synapses decreases inversely to $n_i$. A reduced number of incoming synapses is good for efficient hardware implementation, yet the synaptic activity is a more significant marker for theoretical energy consumption. As depicted in the third plot of Fig.~\ref{fig:compar_PokerDVS}, the number of synaptic events is significantly lower for the line-based preprocessing, neighbouring the ``no preprocesssing'' and up to 2 orders of magnitude smaller than the values obtained for FC layer and convolution. Those different observations combined lead us to the conclusion validated by the fourth plot in Fig.~\ref{fig:compar_PokerDVS}: the line-based preprocessing is the only approach able to reach an inference efficiency similar or better than the one obtained by the baseline embodied by the ``no-preprocessing''. Both convolution and FC layer approaches reach an inference efficiency 1 to 2 orders of magnitude lower than the baseline; although the convolution interestingly stagnates around the same efficiency value, thus adequately balancing accuracy and synaptic activity for a varying $n_i$. \newline

As both convolution and deep fully-connected networks reach significantly lower inference efficiency, the metric targeted in this work to assess the theoretical energy consumption, on the simple task of classifying the PokerDVS dataset, we definitely discard those two approaches in the remainder of this paper despite their demonstrated potential in neuromorphic vision.

\section{Experimental performance on benchmark datasets}
The previous section describes the initial results we obtained on PokerDVS. We validate the line-based preprocessing approach for reducing theoretical energy efficiency on a toy dataset allowing for simple classification of native neuromorphic data. The following section now evaluates the impact of our approach on two more complex benchmark datasets, also natively neuromorphic and widely used for visual classification.

\subsection{Large-scale dataset}
We first extend our study to the Neuromorphic-MNIST (N-MNIST) dataset~\cite{NMNIST}. Recorded by presenting handwritten digits from 0 to 9 to a DVS camera of size $34 {\times} 34$ pixels moving in defined saccades, this dataset comprises more than \qty{70,000}{} samples evenly distributed over 10 classes. It was created to showcase the advantages its small resolution brings, namely the ``rapid testing and iteration of algorithms when prototyping new ideas''~\cite{NMNIST}. Fig.~\ref{fig:output_NMNIST_6_2d_split} illustrates the output of one line-based preprocessing strategy (namely ``corner quarters - inner detectors'') applied to one N-MNIST class. 

\begin{figure}[ht]
    \centering
    \includegraphics[trim={0 0 0 1.1cm},clip,width=\textwidth]{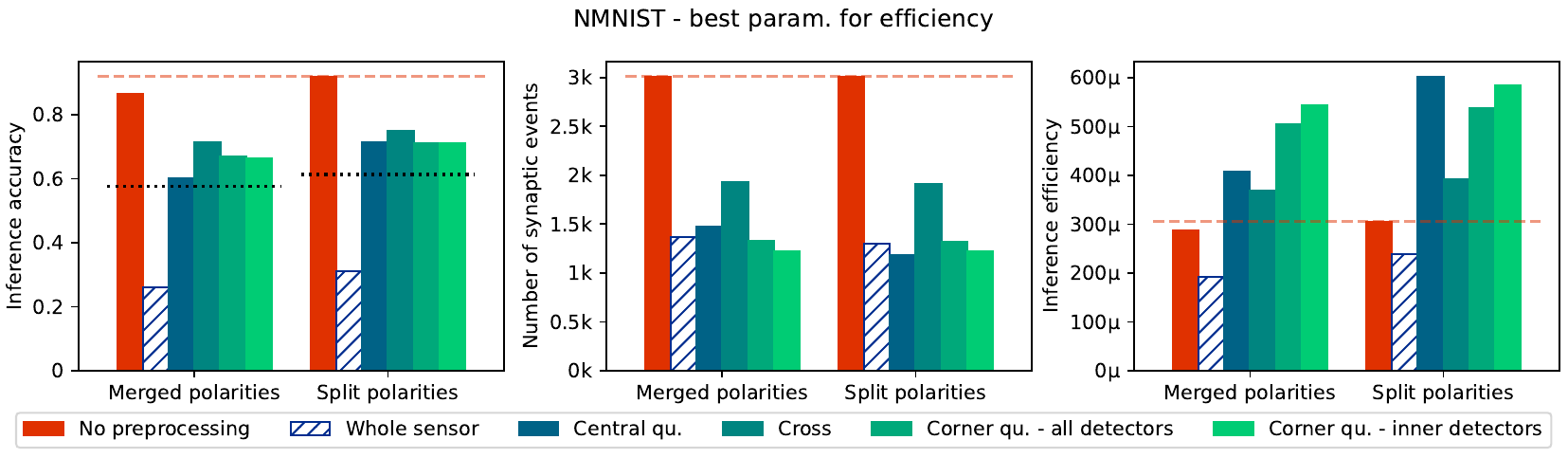}
    \caption{Comparative performance on N-MNIST of the inference accuracy (left), the synaptic activity at inference (middle) and the inference activity (right) of the five line preprocessing approaches with the ``no preprocessing'', distinguished according to polarity handling (merged versus split). The line-based preprocessing strategies' results correspond to the ``best efficiency parametrisation''. The dashed red line indicates the highest value reached by the ``no preprocessing''; the dotted black line (in the first plot) indicates the $\mathds{A}_\text{T}$ threshold. The ``whole sensor'' strategy is hashed out as it does not meet the efficiency condition detailed in Eq.~\ref{eq:threshold}, i.e. its accuracy is lower than two thirds of the ``no preprocessing'' one.}
    \label{fig:effNMNIST}
\end{figure}

Fig.~\ref{fig:effNMNIST} presents the inference accuracy and efficiency of the classification of N-MNIST with and without the application of the various line-based preprocessing strategies introduced in this work, as well as the corresponding number of synaptic events generated over the whole network. These results are achieved using the ``best efficiency parametrisation''. Preprocessing has a stronger impact on N-MNIST classification performance than on PokerDVS': the ``cross'' strategy with split polarities leads to the higher classification accuracy out of the five strategies studied in this work (with ``best efficiency parametrisation'') -- but its value equals only \qty{0.748}{}, i.e. only \qty{81.4}{\%} of the ``no preprocessing'' one. Nevertheless, it achieves this result with only \qty{1905}{} synaptic events, i.e. two third of the number required by the ``no preprocessing'' -- even though the network involved in preprocessing contains an additional intermediate layer. A similar behaviour is observed in all preprocessing strategies, with the exception of ``whole sensor'': all four lead to an inference efficiency exceeding the ``no preprocessing'' one, with the ``central quarter'' strategy with split polarities doubling this baseline value. 

The ``whole sensor'' value is displayed here (albeit hashed out) despite the fact that it does not meet the efficiency condition detailed in Eq.~\ref{eq:threshold}, i.e. its accuracy is lower than two thirds of the ``no preprocessing'' one. Its low accuracy is combined with a reduced number of synaptic events, however they do not balance each other out to produce a good enough trade-off; it is the only strategy not overcoming ``no preprocessing'' in terms of inference efficiency when applied on N-MNIST. It is interesting to note that its low accuracy cannot solely be explained by the low number of synaptic events, but probably rather by a poor spatial density linked to the synaptic patterns, as ``central quarter'' and both ``corner quarters'' strategies have similar or lower synaptic activity and lead to an accuracy twice as high. This underlines the need for relevant synaptic patterns within event data preprocessing for low energy consumption, as this behaviour reverberates positively on efficiency. \newline

All in all, the preprocessing approach previously assessed on a simple dataset effectively leads to a lower theoretical energy consumption for larger-scale neuromorphic classification. 

\subsection{Real-life scenario}

Both neuromorphic datasets evaluated previously share one distinctive characteristic: events are generated from still images (representing card pips for PokerDVS~\cite{PokerDVS} and handwritten digits for N-MNIST~\cite{NMNIST}) moving over uniform backgrounds. Albeit natively neuromorphic, these datasets do not fully showcase the added-value of using an event camera for the detection of fine movements. The following section studies the application of line-based preprocessing on a neuromorphic dataset avoiding this defect and specifically assessing temporal information: the DVS128 Gesture dataset~\cite{amir2017low}. Targeting hand gesture recognition, it is composed of more than 130, 6-second long samples representing around 30 human subjects under different lightning conditions. The subjects moved their hands to match 11 different motions in front of a DVS128 camera. As N-MNIST and PokerDVS contain mainly spatial information, only \qty{10}{\milli\second} of each sample was required for classification (see Tab.~\ref{tab:datasets}); however, a classifier requires at least \qty{1}{\second} per sample to accumulate enough temporal information in order to recognise a hand gesture from DVS128 gesture. 

\begin{figure}[ht]
    \centering
    \includegraphics[trim={0 0 0 1.1cm},clip,width=\textwidth]{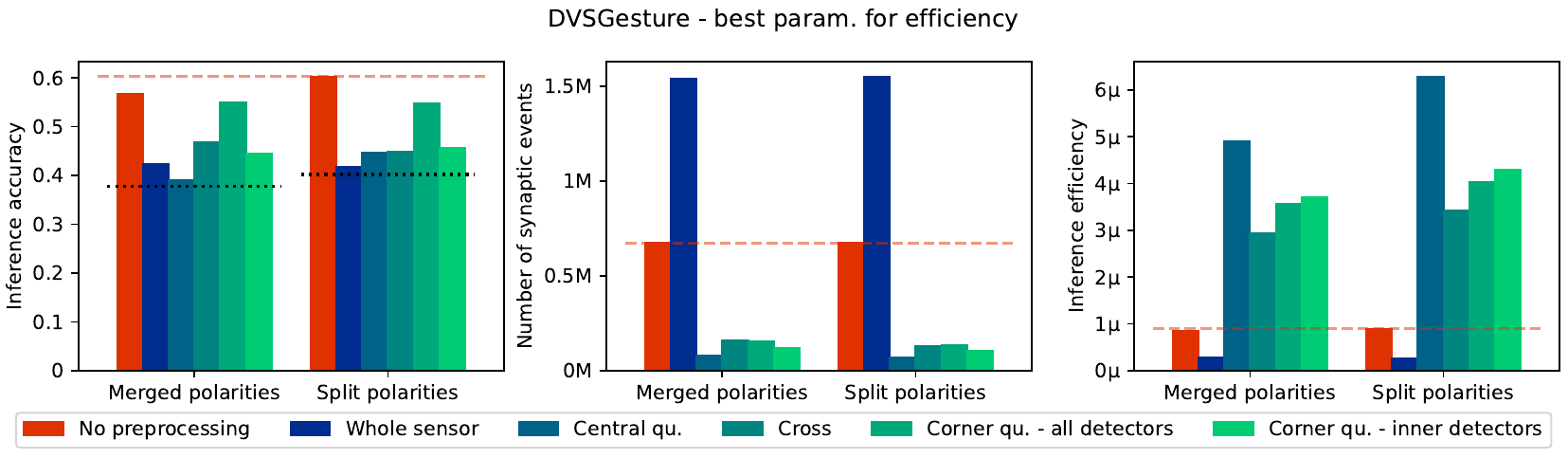}
    \caption{Comparative performance on DVS128 Gesture of the inference accuracy (left), the synaptic activity at inference (middle) and the inference efficiency (right) of the five line preprocessing approaches with the ``no preprocessing'', distinguished according to polarity handling. The line-based preprocessing strategies' results correspond to the ``best efficiency parametrisation''. The dashed red line indicates the higher value reached by the ``no preprocessing''; the dotted black line (in the first plot) indicates the $\mathds{A}_\text{T}$ threshold.}
    \label{fig:effDVSG}
\end{figure}

Fig.~\ref{fig:effDVSG} presents the inference accuracy and efficiency of DVS128 Gesture classification with or without application of the five line-based preprocessing strategies with ``best efficiency parametrisation'', as well as the corresponding number of synaptic events. Similarly as N-MNIST, the ``whole sensor'' approach struggles to reach an optimal trade-off relevant for low energy consumption: its accuracy only reaches \qty{70}{\%} of the ``no preprocessing'' one whereas the corresponding synaptic activity nearly triples, leading to low efficiency. The four other strategies however significantly overcomes the ``no preprocessing'' in terms of theoretical energy consumption, as their inference efficiency is at least 3 times higher (``cross'' with merged polarities), and at best more than 6 times higher (``central quarter'' with split polarities''). This can be explained by the remarkably low number of synaptic events generated, requiring only \qty{16.51}{\%} (split polarities) to \qty{19.05}{\%} (merged polarities) of the synaptic activity of ``no preprocessing'', while reaching up to \qty{96}{\%} of the baseline classification accuracy (``corner quarters - all detectors'' with merged polarities). 

\begin{figure}[ht]
    \centering
    \includegraphics[width=\textwidth]{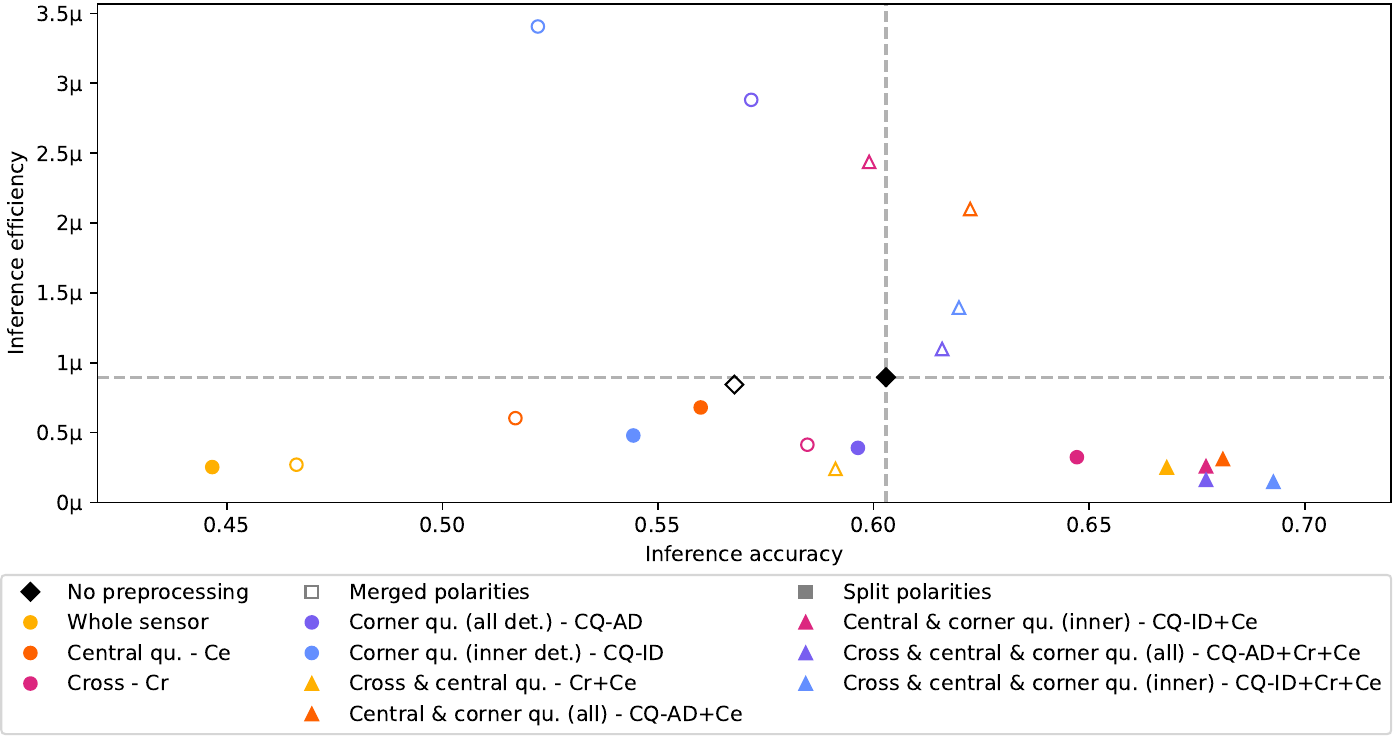}
    \caption{Comparative performance on DVS128 Gesture of the trade-off between efficiency (y-axis) and accuracy (x-axis) of the various approaches obtained at inference for the ``best accuracy parametrisation'' (as opposed to the ``best efficiency parametrisation'' presented in Fig.~\ref{fig:effDVSG}). Are pictured the ``no preprocessing'' as well as the five line-based preprocessing introduced in this work; are also pictured five cumulative approaches, i.e. combinations of two or three of the initial line-based preprocessing strategies. The results with merged polarities are identified with unfilled markers, the ones with split polarities with filled markers. The vertical and horizontal dashed black lines indicate respectively the higher accuracy and efficiency values reached by the ``no preprocessing''. All strategies meet the $\mathds{A}_\text{T}$ condition, i.e. obtain accuracies higher than two thirds of the original.}
    \label{fig:accDVSG}
\end{figure}

As DVS128 Gesture carries more complex information, our line-based preprocessing strategies might not be complex enough to effectively capture all relevant features for classification, thus leading to a not-ideal performance. We introduce here a final variant of our line-based preprocessing: five cumulative strategies composed of two to three strategies combined (namely ``central quarter'' added to ``cross'' and to both ``corner quarters'' as well as the cumulation of all three) to refine features extraction. In the remainder of this section, they will be referred to with shortened names where Ce corresponds to ``central'', Cr to ``cross'', and CQ-AD and CQ-ID to ``central quarters -- all detectors'' and ``inner detectors'' respectively, with a symbol + denoting the cumulation. The synapses and neural populations of the initial strategies and the generated events are summed together, then fed to the classifier. Fig.~\ref{fig:accDVSG} presents the trade-off between accuracy (x-axis) and efficiency (y-axis) results obtained using ``no preprocessing'' (in red), the five initial line-based preprocessing strategies and the five cumulative ones, for both training and inference with ``best accuracy parametrisation''. The ``no preprocessing'' sets the baseline we aim to overcome: outperforming either its accuracy (vertical dashed red line) or its efficiency (horizontal dashed red line), where the upper right corner of the plot corresponds to the ideal scenario of outperforming both simultaneously. 

According to Fig.~\ref{fig:accDVSG}, this ideal scenario is indeed achieved at inference thanks to the cumulative strategies: Cr+Ce, CQ-AD+Cr+Ce and CQ-ID+Cr+Ce all three with merged polarities reach an higher accuracy than ``no preprocessing'' with an increased efficiency -- Cr+Ce more than doubling the baseline efficiency. Additionally, multiple approaches outperforms the baseline set by ``no preprocessing'' in terms of pure classification accuracy: most cumulative approaches do, but can also be mentioned the ``cross'' approach with split polarities which outperforms by 5 points the baseline. Finally, it is quite interesting to note that cumulative strategies with merged or split polarities tend to respectively lead to greater efficiency or greater accuracy than ``no preprocessing''. As theorised previously, this can be explained by the fact that considering the polarities separately retains more temporal information and finer details, allowing for a more precise classification and a higher accuracy, especially for a ``best accuracy parametrisation'' which optimises the latter. In contrary, merging polarities leads to lost information and a decreased classification accuracy -- especially for DVS128 Gesture, where the direction of the hand gesture can be partially inferred from the spatial distribution of the events polarities -- while decreasing the number of synaptic events, thus increasing the efficiency. \newline 

All in all, these last experimental results demonstrate that our line-based preprocessing allows for a more efficient and, in specific cases, a better performing neuromorphic classification of complex visual data corresponding to real-life scenarios.

\section{Discussion}
As stated in the introduction of this paper, we aim to improve the energy efficiency in a traditional neuromorphic computer vision task; this goal is to be achieved while maintaining a satisfying performance, meaning that improving the latter would be a secondary benefit and not a mandatory objective to achieve. Five main strategies for line-based preprocessing are evaluated for this purpose on three datasets: PokerDVS~\cite{PokerDVS}, N-MNIST~\cite{NMNIST} and DVS128 Gesture~\cite{amir2017low}. The latter is also submitted to a cumulation of strategies in order to capture finer details of this more complex dataset. 

Tab.~\ref{tab:sumup_results} summarises the main results of our work, by highlighting the best strategies for the classification of the three event-based datasets while optimising either classification accuracy or theoretical energy efficiency. According to Tab.~\ref{tab:sumup_results}, the ``best accuracy parametrisation'' leads to an accuracy satisfyingly close to the baseline set by ``no preprocessing'' for PokerDVS and N-MNIST. In the case of DVS128 gesture, the cumulation of multiple strategies allows for an increase of the accuracy by 6 to 9 points compared to the baseline without preprocessing. Furthermore, the ``best efficiency parametrisation'' systematically leads to an increased efficiency while conserving a satisfying accuracy (at least three quarters of the baseline one, above the two thirds limit set by $\mathds{A}_\text{T}$ in Eq.~\ref{eq:threshold}). The efficiency of PokerDVS is multiplied by 2.5, the one of N-MNIST is doubled and best of all, the efficiency of DVS128 Gesture is multiplied by 7. Finally, the strategies cumulation applied to the last scenario also reaches a trade-off increasing significantly both accuracy and efficiency.

\begin{table}[h]
    \caption{Classification performance in terms of accuracy and efficiency of the ``no preprocessing'' baseline and the line-based preprocessing strategies leading to best accuracy and best efficiency, for the three event-based datasets PokerDVS, N-MNIST and DVS128 Gesture. The values in bold correspond to the best accuracy and best efficiency obtained for each dataset.}
    \centering
    \small
    \begin{tabular}{l | c c | c c c | c c }
        \hline
        \textbf{Dataset} & Parametrisation & Strategy & Polarities & $k$ & $\omega$ & Accuracy $\mathds{A}$ & Efficiency $\mathds{E}$ \\
        \hline
        \multirow{3}{*}{PokerDVS} & \multicolumn{2}{c |}{No preprocessing} & either & / & / & \textbf{1.0} & \num{2.171e-4} \\
         & Best accuracy & CQ-AD & merged & 20 & 7.5 & \textbf{1.0} & \num{1.601e-4} \\
         & Best efficiency & Ce & merged & 30 & 5 & 0.958 & \textbf{\num{5.43e-4}} \\
        \hline
        \multirow{3}{*}{N-MNIST} & \multicolumn{2}{c |}{No preprocessing} & split & / & / & \textbf{0.919} & \num{3.054e-4} \\
         & Best accuracy & Cr & split & 30 & 10 & 0.901 & \num{1.935e-4} \\
         & Best efficiency & Ce & split & 30 & 7.5 & 0.713 & \textbf{\num{6.027e-4}} \\
        \hline
        \multirow{4}{*}{\begin{tabular}{c}DVS128 \\ Gesture \\\end{tabular}} & \multicolumn{2}{c |}{No preprocessing} & split & / & / & 0.603 & \num{8.966e-7} \\
         & Best accuracy & CQ-ID+Cr+Ce & split & 1 & 7.5 & \textbf{0.693} & \num{1.479e-7} \\
         & Best efficiency & Ce & split & 25 & 10 & 0.447 &\textbf{\num{62.89e-7}} \\
         & \textbf{\textcolor{teal}{Best overall}} & CQ-AD+Ce & merged & 15 & 7.5 & \textbf{\textcolor{teal}{0.622}} & \textbf{\textcolor{teal}{\num{21.00e-7}}} \\
        \hline
    \end{tabular}
    \vspace{0.2cm} \\
    \footnotesize
    \textit{Note: Ce stands for the ``central quarter'' strategy, Cr for ``cross'', CQ-AD for ``corner quarters - all detectors'' and CQ-ID for ``corner quarters - inner detectors''. A symbol + denotes the cumulation of multiple strategies together.}
    \label{tab:sumup_results}
\end{table}

Thanks to Tab.~\ref{tab:sumup_results}, we thereby confirm our initial hypothesis of an increased theoretical energy efficiency thanks to the application of relevant event preprocessing to neuromorphic classification, on three datasets varying in ease, data properties and scale. In addition, we demonstrate that complex scenarios may gain from line-based preprocessing both in accuracy and in efficiency, as highlighted by the ``Best overall'' results on DVS128 Gesture (last row of Tab.~\ref{tab:sumup_results}). \newline

It is interesting to note that the ``whole sensor'' approach performs systematically poorly. This can be explained by multiple parameters: the line detection mechanism positively detects lines and outputs corresponding events solely when the aforementioned lines cross the borders of the monitored sensor region. Most event-based datasets for classification, and particularly those chosen in this work, carry centralised information, i.e. the important events are distributed in the center of the sensor. As the ``whole sensor'' approach encompass the whole sensor (as its name suggests), only few lines contained in the centralised event data do cross the borders of its monitored region. This approach thus requires a high activation (i.e. a high strength $\omega$) to produce events, and those events should be numerous to carry relevant information. ``Whole sensor'' therefore either:
\begin{itemize}
    \item does not produce events, as is the case for PokerDVS; 
    \item does not reach a good enough accuracy to be considered in our work, as is the case for N-MNIST (see Fig.~\ref{fig:effNMNIST}) -- as a reminder, we only study the efficiency of strategies reaching at least two thirds of the baseline accuracy;
    \item reaches a satisfying performance by outputting an overwhelmingly large number of synaptic events, thus leading to a significantly lower efficiency, as is the case for DVS128 Gesture (see Fig.~\ref{fig:effDVSG}).
\end{itemize}

Additionally, the ``whole sensor'' approach relies on 4 detectors, i.e. at least twice less as many as the other approaches (except the ``central quarter'' one); this small number does not allow for fine enough details to be captured, thus for a good enough classification. ``Central quarters'' overcomes this issue by monitoring a smaller region of the sensor, where relevant information is more susceptible to be perceived by the line detection model.\newline

We can observe in both Fig.~\ref{fig:effNMNIST} (right) and Fig.~\ref{fig:effDVSG} (right) that strategies applied to N-MNIST and DVS128 Gesture with split polarities perform better overall than their counterpart; this can be explained due to the loss of relevant information provided by the polarities, which unbalances the benefits provided by the lower number of synaptic connections within the network (see Fig.~\ref{fig:architecture}).
Such a behaviour is less observed on PokerDVS, as the simplicity of this first dataset allowed for similar performance whether the polarities are merged or split. Fig.~\ref{fig:hyperparam_PokerDVS} shows similar tendencies for each preprocessing strategy for both merged (top row) and split (bottom row) polarities. \newline

Finally, this work do not intend to increase classification accuracy, but rather to validate the energy improvement brought by event preprocessing. Nonetheless, we do observe a significant increase in classification accuracy on DVS128 Gesture thanks to event preprocessing (see Tab.~\ref{tab:sumup_results}). Those results are still far from those obtained in the recent literature: state-of-the-art neuromorphic classification surpasses 95\% accuracy on this dataset~\cite{shi2025spikingbraincompressionexploring,Abdennadher2025,malettira2025tskipsefficiencyexplicittemporal}. However, the latter are obtained using more complex architectures than the deliberately simple one studied in this work ; such observations therefore point towards promising future classification performances thanks to preprocessing.

More strikingly, the ``best overall'' behaviour observed on DVS128 Gesture (see Tab.~\ref{tab:sumup_results}, last row) stands out as a successful trade-off between classification performance and energy consumption, with a simultaneous increase in both accuracy and efficiency. Now that the positive impact of line-based event preprocessing on energy efficiency has been demonstrated in the context of a deliberately simple SNN classifier, the road is open for its application on more complex models to reproduce such ``best overall'' behaviour and conjointly optimise energy and performance.

\section{Conclusion}
To the best of the authors' knowledge, this work introduces the first end-to-end neuromorphic line-based preprocessing mechanism ever applied to event data. We evaluate the positive impacts on theoretical energy efficiency of such an approach applied to the neuromorphic classification of several event-based datasets varying in ease, data properties and scale. 

We demonstrate the multiple advantages of preprocessing event data before performing a neuromorphic computer vision task. We highlight improvements in terms of:
\begin{itemize}
    \item energy efficiency: preprocessing significantly and systematically reduces the number of synaptic events while maintaining classification accuracy for a tuned set of hyperparameters $(k,\omega)$, reaching an energy efficiency up to 7 times greater than the baseline in a complex classification task;
    \item memory usage: line-based preprocessing systematically requires a lower number of incoming synapses per neuron, which is one of the main challenges in the hardware implementation of neuromorphic architectures -- and not the number of neurons and synapses per se, thanks to time-division multiplexing;
    \item performance: applying line-based preprocessing in complex, real-life scenarios as depicted in DVS128 Gesture leads to a 9-point increase of the baseline accuracy. 
\end{itemize}

Finally, our more complex scenario optimises simultaneously both performance and theoretical energy consumption \textit{via} the cumulation of multiple strategies.
Our results lay down the groundwork for event preprocessing and highlight its necessity for an efficient, low-energy neuromorphic computer vision. Similarly to traditional computer vision involving deep ANNs and standard cameras, preprocessing must indeed be applied to the input data before any downstream handling. Furthermore, preprocessing must be adequate and adaptable to the considered data and the tasks undertaken to reverberate positively on low energy consumption. \newline

Future works include further exploration of the results presented in the core of this paper, over a larger set of computer vision tasks involving larger amount of data, such as segmentation on the Prophesee 1 Megapixel Dataset~\cite{one_megapixel}. We wish to further explore the simultaneous optimisation of accuracy and efficiency observed in DVS128 Gesture, with more complex and/or realistic tasks. Additionally, we wish to study in greater depth the influence of other neuronal and synaptic parameters on the preprocessing. Classification performance could also be assessed using a more bio-inspired and hardware-friendly learning strategy, giving more importance to the temporal aspect, such as~\cite{lewden2023}. The study of theoretical energy consumption, approximated by the synaptic activity in this work, could be deepened with more complex model of neuromorphic energy consumption such as~\cite{dampfhoffer2023,lemaire2022ICONIP}. Finally, as results presented above were obtained using a CPU simulator~\cite{gewaltig_nest_2007}, these preprocessing mechanisms are to be deployed on neuromorphic hardware such as SpiNNaker~\cite{spinnaker} or on FPGAs in order to assess its application in real-time and actual energy consumption, thus validating our promising, theoretical energy gain.

\section*{Acknowledgments}
This work is supported by a public grant overseen by the French National Research Agency (ANR) as part of the ‘PEPR IA France 2030’ programme (Emergences project ANR-23-PEIA-0002). 

\section*{Authors' contributions}
The authors Amélie Gruel, Pierre Lewden, Adrien F. Vincent and Sylvain Saïghi contributed to the conceptualisation, methodology design and formal analysis of the study. The project coordination, data curation, investigation, software implementation, validation and visualisation were handled by Amélie Gruel. Sylvain Saïghi and Adrien F. Vincent carried out the funding acquisition. The first draft of the manuscript was written by Amélie Gruel; Pierre Lewden, Adrien F. Vincent and Sylvain Saïghi added to a second draft by reviewing and editing. All authors read and approved the final manuscript.

\bibliographystyle{ieeetr}
\bibliography{biblio}

\end{document}